\documentclass{osa-article}

\journal{oe}
\usepackage{url}
\usepackage{amsmath}
\usepackage{amssymb}
\usepackage{times}
\usepackage{epsfig}
\usepackage{graphicx}
\usepackage{multirow}
\usepackage{diagbox}
\usepackage{threeparttable}
\usepackage{cite}
\usepackage{url}
\usepackage[ruled]{algorithm2e}
\usepackage{gensymb}
\usepackage{float}
\usepackage{lipsum} % Only used to generate random text.

\begin{document}

\title{Agile wide-field imaging with selective high resolution}

\author{Lintao Peng,\authormark{1} Liheng bian,\authormark{1,*} Tiexin Liu\authormark{1} and Jun Zhang\authormark{1}}

\address{\authormark{1}School of Information and Electronics \& Advanced Research Institute of Multidisciplinary Science, Beijing Institute of Technology, Beijing 100081, China}

\email{\authormark{*}bian@bit.edu.cn} %% email address is required



\begin{abstract}
Wide-field and high-resolution (HR) imaging are essential for various applications such as aviation reconnaissance, topographic mapping, and safety monitoring. The existing techniques require a large-scale detector array to capture HR images of the whole field, resulting in the high complexity and heavy cost. In this work, we report an agile wide-field imaging framework with selective high resolution that requires only two detectors. It builds on the statistical sparsity prior of natural scenes that the important targets locate only at small regions of interest (ROI), instead of the whole field. Under this assumption, we use a short-focal camera to image a wide field with a certain low resolution and use a long-focal camera to acquire the HR images of ROI. To automatically locate ROI in the wide field in real-time, we propose an efficient deep-learning-based multiscale registration method that is robust and blind to the large setting differences (focal, white balance, etc) between the two cameras. Using the registered location, the long-focal camera mounted on a gimbal enables real-time tracking of the ROI for continuous HR imaging. We demonstrated the novel imaging framework by building a proof-of-concept setup with only 1181 gram weight and assembled it on an unmanned aerial vehicle for air-to-ground monitoring. Experiments show that the setup maintains 120$^{\circ}$ wide field-of-view (FOV) with selective 0.45$mrad$ instantaneous FOV.
\end{abstract}

%%%%%%%%%%%%%%%%%%%%%%%%%%  body  %%%%%%%%%%%%%%%%%%%%%%%%%%
\section{Introduction}
 {W}{ide}-field and high-resolution (HR) imaging is essential for various applications such as aviation reconnaissance, topographic mapping and safety monitoring \cite{romeo2019use,law2016evryscope,eschmann2012unmanned,heymsfield2015implementing}. Unfortunately, the field of view (FOV) and resolution of a camera are usually limited to each other, due to the fixed system throughput \cite{zheng2013wide}. For the cameras with long focal lengths, the resolution is high but the field of view is narrow, and vice versa. The existing solution to the contradiction is to build a camera array system, and stitch the HR images of each camera together to extend the field of view\cite{multiscale,development,yuan2017multiscale,cossairt2011gigapixel,kopf2007capturing,cossairt2016camera,golish2012development}. The hardware is complex with high cost and heavy electrical complexity, and the data amount is huge that is difficult to transmit and process in real-time. In addition, it is hard to optionally adjust focal length simultaneously for all the cameras. 

\begin{figure}
	\centering
	\includegraphics[scale=0.5]{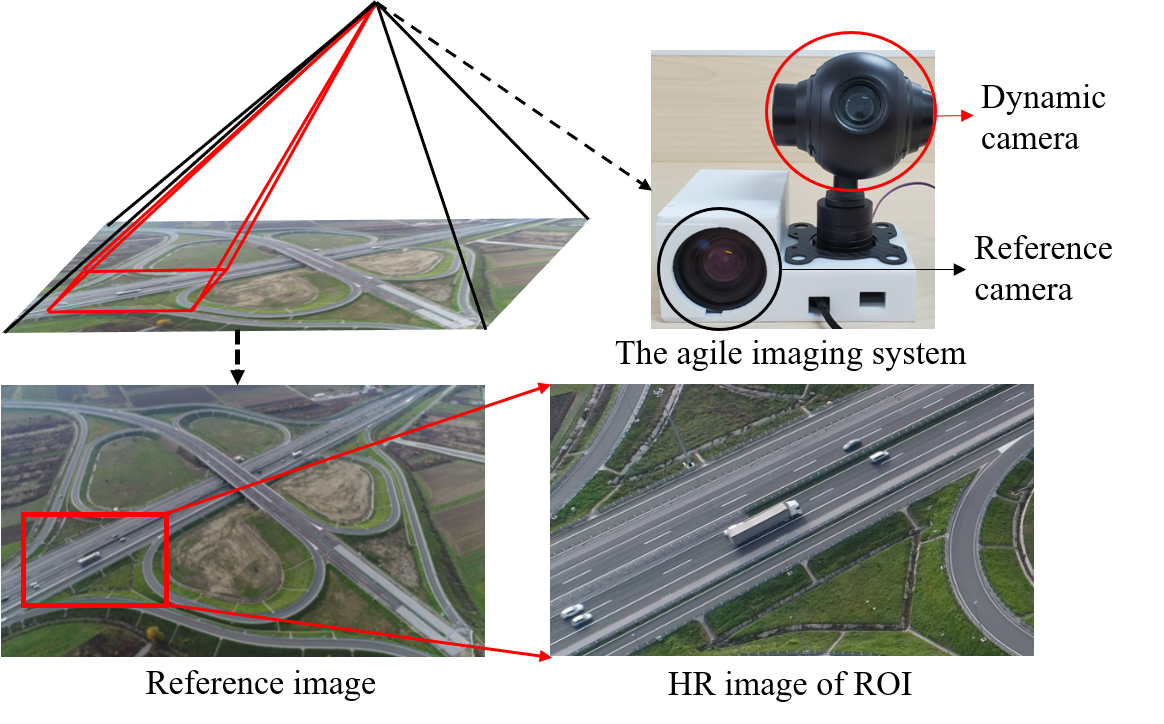}
	\caption{The reported agile wide-field imaging framework with selective high resolution. It is constructed of a reference camera to acquire wide-field images, and a dynamic camera to automatically and adaptively track regions of interest and acquire high-resolution images. The entire system has only 1181 gram weight, with 120$^{\circ}$ wide field-of-view (FOV) with selective 0.45$mrad$ instantaneous FOV.}
	\label{figure:1}
\end{figure}

In practical applications, however, the regions of interests (ROI) occupy only a small part of the wide FOV, known as the statistical sparsity prior in the spatial domain \cite{ren2018small}. Taking aerial reconnaissance as an example, as shown in Fig. \ref{figure:1}, the important target (cars) is much smaller compared to the entire field of view. In other words, there is no need to acquire HR information of all the subfields that is laboursome. Instead, HR imaging of only the key ROI provides sufficient information for most applications.

\begin{figure*}[t]
	\centering
	\includegraphics[width=1.1\linewidth]{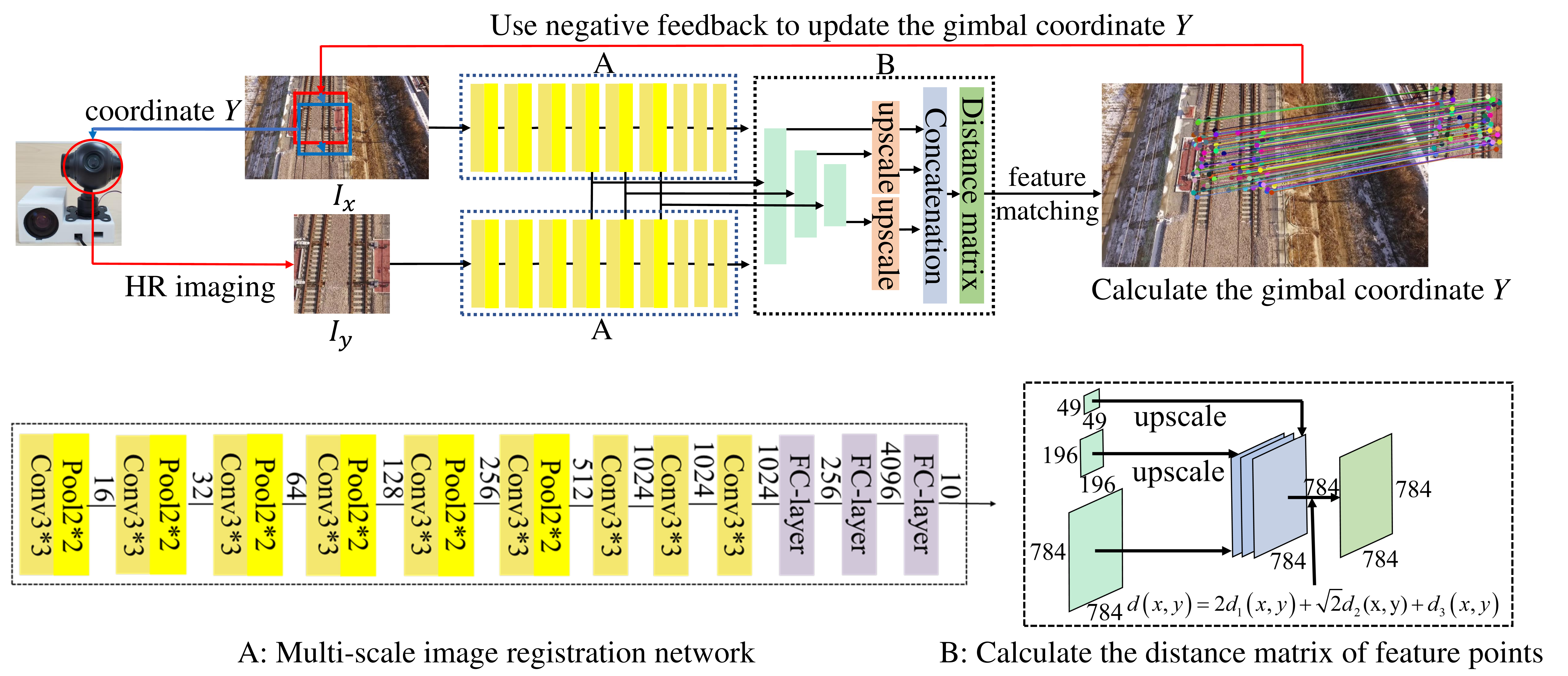}
	\caption{The workflow of the reported wide-field and high-resolution imaging technique. The reference camera is employed to take wide-field images $I_x$ with a certain low resolution, and the dynamic camera is to acquire HR images $I_y$ of ROI. A deep-learning-based algorithm is employed to register the different-scale images, in order to find the location relationship between the two fields of views of the two cameras. The algorithm first extracts multi-scale feature descriptors from the output of multiple different intermediate network layers  and then calculates the feature point distance matrix based on these feature descriptors. Under a bidirectional feature point matching strategy, high-accuracy feature point matching pairs are obtained that produce the homography transformation matrix of the two input images. The coordinate $Y$ of $I_y$ in $I_x$ is calculated using the homography transformation matrix, and input to the gimbal to update the field of view. By the negative feedback mechanism, the entire system enables continuous wide-field and HR imaging.}
	\label{figure:2}
\end{figure*}

Based on the above observations, we report an agile wide-field imaging framework with selective high resolution, as shown in Fig. \ref{figure:1} and Fig. \ref{figure:2}. The system requires only two cameras, including a short-focal camera to image wide field for reference, and a long-focal camera mounted on a gimbal\cite{knowles2008solid} to acquire HR images of ROI. To automatically locate ROI in the wide field, we propose an efficient deep-learning-based multiscale registration method that enables real-time registration of the HR image with the wide-field low-resolution (LR) image. The technique is robust and blind to the large setting differences (focal length, white balance, etc) between the two cameras, while the conventional registration algorithms \cite{cossairt2011gigapixel,philip2014distributed,yuan2017multiscale,son2011multiscale} fail in such a case that different devices maintain different unknown imaging parameters. In such a multiscale registration strategy, the long-focal camera mounted on a gimbal enables real-time tracking of the ROI for continuous HR imaging, and the system is robust to platform vibration and fast target movement with negative feedback. 

In summary, the main innovations of the reported agile imaging framework include:
\begin{itemize}
	\item The system is agile with only two cameras compared to the conventional large-scale detector arrays. Together with a gimbal, the system enables continuous HR imaging of ROI in wide-field, with a convenient focal adjustment that is not required to be known. The multiscale imaging strategy also ensures strong robustness to platform vibration and fast target movement.
	\item We report an efficient multiscale registration technique to automatically locate ROI in the wide field. By fusing multiscale convolution features to construct multiscale feature descriptors, the technique is robust and blind to different imaging parameters of the two cameras and maintains real-time running efficiency (0.1s/frame).
	\item We built a prototype setup that integrates a JAI GO-5000 camera and a Foxtech Seeker10 camera for imaging, and an NVIDIA TX2 processor for real-time processing and control. The setup's weight is only 1181 grams, making it applicable for load-limited platforms such as an unmanned aerial vehicle (UAV). Experiments validate its 120$\degree$ wide FOV with selective 0.45$mrad$ instantaneous FOV.
\end{itemize}

The rest of this article is organized as follows. The details of the reported imaging framework are presented in section 2. The experiment results are shown in section 3. In section 4, we conclude this work with further discussions.

\section{Method}

%图2

\subsection{The agile imaging framework}
The reported agile wide-field and high-resolution imaging system is shown in Fig. \ref{figure:1}, which consists of a reference camera with a short focal length, and a dynamic camera with a long focal length. The reference camera is fixed to acquire wide-field images but with low resolution. The dynamic camera is mounted on a gimbal, which is rotated to track ROI and acquire its HR images. 

To realize robust tracking, the location of the narrow-field HR images (ROI) on the wide-field image should be identified in real-time, which can be realized using the following introduced multiscale registration technique. Then, we convert the location of the ROI in the reference camera into coordinate $Y$ of the dynamic HR camera. The coordinate conversion relationship between the static reference camera and the dynamic HR camera is calibrated in advance. Afterward, the dynamic HR camera is rotated to the coordinate $Y$ to take HR images. The HR images are further matched to the wide-filed LR images to verify whether the coordinates of the dynamic HR camera rotation are correct. By iteratively updating the coordinate in a negative feedback manner as shown in Fig. \ref{figure:2}, the coordinate conversion relationship between the cameras can be adaptively corrected, and continuous tracking and HR imaging of ROI is realized.

\subsection{Blind multiscale registration}

To locate ROI in the wide field for continuous tracking and HR imaging, we propose an efficient multiscale registration algorithm enabling to register $I_x$ and $I_y$ captured by the two cameras. Due to the large focal difference between the two cameras, there exists a large scale difference between $I_x$ and $I_y$. Besides, the focal difference (i.e., scale difference) is usually unknown in practical applications. What's more, the two cameras may maintain different white balance settings, resulting in different spectrums of the two images. In such a case, the conventional image registration methods fail to find correspondence between these two images.

Considering that the agile imaging system requires real-time efficiency in practical applications, we employ deep learning for blind multiscale registration. Briefly, the feature points and corresponding feature descriptors of the two images are first detected and constructed using a network. Then, they are matched in a bidirectional manner to produce a high-accuracy homography transformation matrix $H$. Finally, the input image $I_y$ is aligned with $I_x$, and the location relationship is obtained. The detailed steps are summarized in Fig. \ref{figure:2}.

\subsubsection{Construct feature descriptor}

To quickly and efficiently extract multi-scale feature maps from the input image, we designed a special multi-scale feature extraction network,  The detailed network structure is shown in Fig. \ref{figure:2}. It has a total of 18 layers, consisting of 9 convolutional layers, 6 max-pooling layers, and 3 fully connected layers, all convolutional layers use 3$\times$3 convolution kernels, and the size of the pooling layer is 2$\times$2. Each of the first six convolutional blocks halves the feature map size and doubles the number of channels. We train our multi-scale feature extraction network on the classification data set CIFAR-10\cite{krizhevsky2009learning} to achieve a classification accuracy of 93\%.  In this way, the network has a strong feature extraction capability.

Based on the visualization results of the convolutional layer\cite{zeiler2014visualizing} and a large number of comparative experiments on extracting the output of different layers, we chose the output of three of the pooling layers to construct our feature descriptors, namely the fourth, fifth, and sixth pooling layer. These layers search for a set of common patterns and generate special characteristic response values, which can cover different sizes of perception domains\cite{coates2011selecting}, making it robust to large unknown scale difference between the input two images. The lower layers are not selected because their output is affected by specific detection objects that are not suitable for detecting general features. Since the network is constructed only by stacking convolutional layers and fully connected layers without using cross-layer connections \cite{he2016deep,DBLP}, it can maintain high operating efficiency. And since we only use convolutional layers to extract features, the input image can be of any size, as long as the height and width are multiples of 32.
However, the size of the input image has two effects:
First, the size of the input network image will affect the image perception domain of each descriptor and affect the performance.
Second, larger input images require more calculations.

Considering comprehensively, we adjust the size of the input image to 448$\times$448, and then propagate it through the network to have an appropriate size of the perceptual domain and reduce calculations. The network generates a feature point for each 16$\times$ 16 pixel block of the input image, we use the middle layer output of the multi-layer convolutional neural network to generate the feature descriptors, including $F1$, $F2$ and $F3$ from the fourth, fifth, and sixth pooling layers. Among them, each $F1$ feature descriptor is generated by 1 feature point, each $F2$ feature descriptor is composed of 4 feature points, and each $F3$ feature descriptor is composed of 16 feature points.

\subsubsection{Calculate the distance matrix of feature points}

As the input image is resized to 448$\times$448 and the feature point extraction block size is 16$\times$16, a total number of 784 128-dimensional $F1$ descriptors, 196 256-dimensional $F2$ feature descriptors, and 49 512-dimensional $F3$ feature descriptors are generated. 
The feature distance between two feature points x and y is a weighted sum of three metrics as
\begin{equation}
d(x,y)=2d_1(x,y)+\sqrt{2}d_2(x,y)+d_3(x,y)\label{eq:distance},
\end{equation}
The weights are employed to balance the different scales of the feature descriptors. Each component distance value is defined as the Euclidean distance between two feature descriptors $D_i(x)$ and $D_i(y)$
%D代表特征描述符，F1类特征描述符的距离d1前面有一个权重2，F2类特征描述符的距离d2前面有一个权重根号2，是因为每个F1类特征描述符为128维，每个F2类特征描述符为256维，每个F3类特征描述符为512维。
\begin{equation}
d_i(x,y)=Euclidean(D_i(x),D_i(y)).
\label{eq:distance_difine}
\end{equation}

The size of the $F1$ feature descriptor of the two images is 784$\times$128, and the size of the feature distance matrix generated by the above calculation is 784$\times$784. To improve the calculation efficiency of the feature distance matrix, we introduce the prior information of the corner coordinates. Specifically, we use the Harris corner detection technique \cite{harris1988combined} to produce the coordinate information of corner points. Then, the coordinate information of the corner points is resized into a 784$\times$1 vector, which is then employed as an indicator of whether the feature points are corner points. Only the corner points are used to calculate the distance between the feature points of the two images. Under the above strategy, the distance matrix of $F2$ (192$\times$192) and $F3$ (49$\times$49) feature descriptors are generated. Finally, the three feature distance matrix are fused together into a characteristic distance matrix (784$\times$784) following Eq. (\ref{eq:distance}).

\subsubsection{Bidirectional feature matching and registration}

To increase feature matching accuracy and improve matching robustness, we employ a bidirectional matching strategy to not only use the feature points in the input image $I_x$ for search but also employ the feature points in the input image $I_y$. As a demonstration, the partial one-way feature point matching algorithm is presented in Algorithm \ref{algo:1}. By denoting two feature points as x and y, and defining a matching threshold $\theta$, we select 128 feature point pairs of the highest matching accuracy. Under this strategy, two sets of feature point pairs are obtained by bidirectional feature point matching. Then, we take the intersection of these two sets as the final feature point matching pairs.

\begin{algorithm}[h]
	
	\caption{The feature point matching algorithm}%算法名字
	\label{algo:1}
	\LinesNumbered %要求显示行号
	\KwIn{Feature point set X,Y extracted from two input images.}%输入参数
	
	Use Eq. (\ref{eq:distance}) to calculate the distance between two feature points\;
	The sorting order in the algorithm is from smallest to largest\; %\;用于换行
	\For{x in point set X}{
		
		\For{y in point set Y}{
			$dis_i=d(x,y)$\;
		}
		$sort(dis)$\;
		$\theta_i=dis_2/dis_1$\;
		
	}
	$\theta_{max}=MAX(\theta)$\;
	$Count=0$\;
	
	\While{$Count<=128$}{
		$\theta_{max}=\theta_{max}-0.01$\;
		\ForEach{$\theta_i$  in  $\theta$}{
			\If{$\theta_i$ $>$ $\theta_{max}$}{
				$Count=Count+1$;
			}
		}
	}
	\KwOut{The matched pairs of feature points with $\theta$ value greater than $\theta_{max}$}%输出
	
\end{algorithm}

Then, the feature point matching pairs are used to calculate the homography transformation matrix\cite{dubrofsky2009homography} between two input images, The homography transformation matrix is a 3$\times$3 matrix, which is calculated in the least square scheme. We employ the RANSC algorithm \cite{fischler1981random} to eliminate poor matching pairs of feature points and make the registration result more accurate. After obtaining the homography matrix, the pixels in one image can be mapped to the corresponding pixels in the other image, and we can finally obtain the coordinate $Y$ for the dynamic camera.

%\cite{dubrofsky2009homography}

\section{Simulation and Experiment}

In the next, we validate the effectiveness of the reported agile imaging framework by both simulation and experiment. We first ran the reported algorithm on the public SUIRD \cite{SUIRD} and OSCD datasets \cite{OSCD} to test its registration matching accuracy. The SUIRD dataset includes 60 pairs of images, which contain viewpoint changes in horizontal, vertical and their mixture with small overlap, image distortion, and severe outliers. The OSCD dataset contains 24 pairs of satellite remote sensing images of the same location taken at different times. As a comparison, we also tested the state-of-the-art image registration techniques, including SIFT\cite{lowe2004distinctive}, ORB\cite{rublee2011orb}, AKAZE\cite{alcantarilla2011fast}, TFeat\cite{balntas2016learning}, HardNet\cite{mishchuk2017working} and Super-point\cite{detone2018superpoint}. The first three are the traditional feature-matching-based registration technique, and the latter three are based on deep learning. Among them, the TFeat and HardNet are both based on the siamese network to generate accurate feature descriptors for image registration, and the Super-point method runs in a self-supervised way to extract feature points and calculate descriptors simultaneously. We also built a proof-of-concept prototype setup and applied it for both ground and air-to-ground monitoring experiments. The details are introduced as follows.

\subsection{Multiscale image registration accuracy}

Considering that the two cameras in the reported agile imaging framework maintain different focal lengths, to validate the effectiveness of the techniques for blind multiscale registration, we synthesized a multiscale copy for each image in both the datasets, with the scale difference being 16, 64, and 256. 

\begin{figure*}[h]
	\centering
	\includegraphics[width=1.0\linewidth]{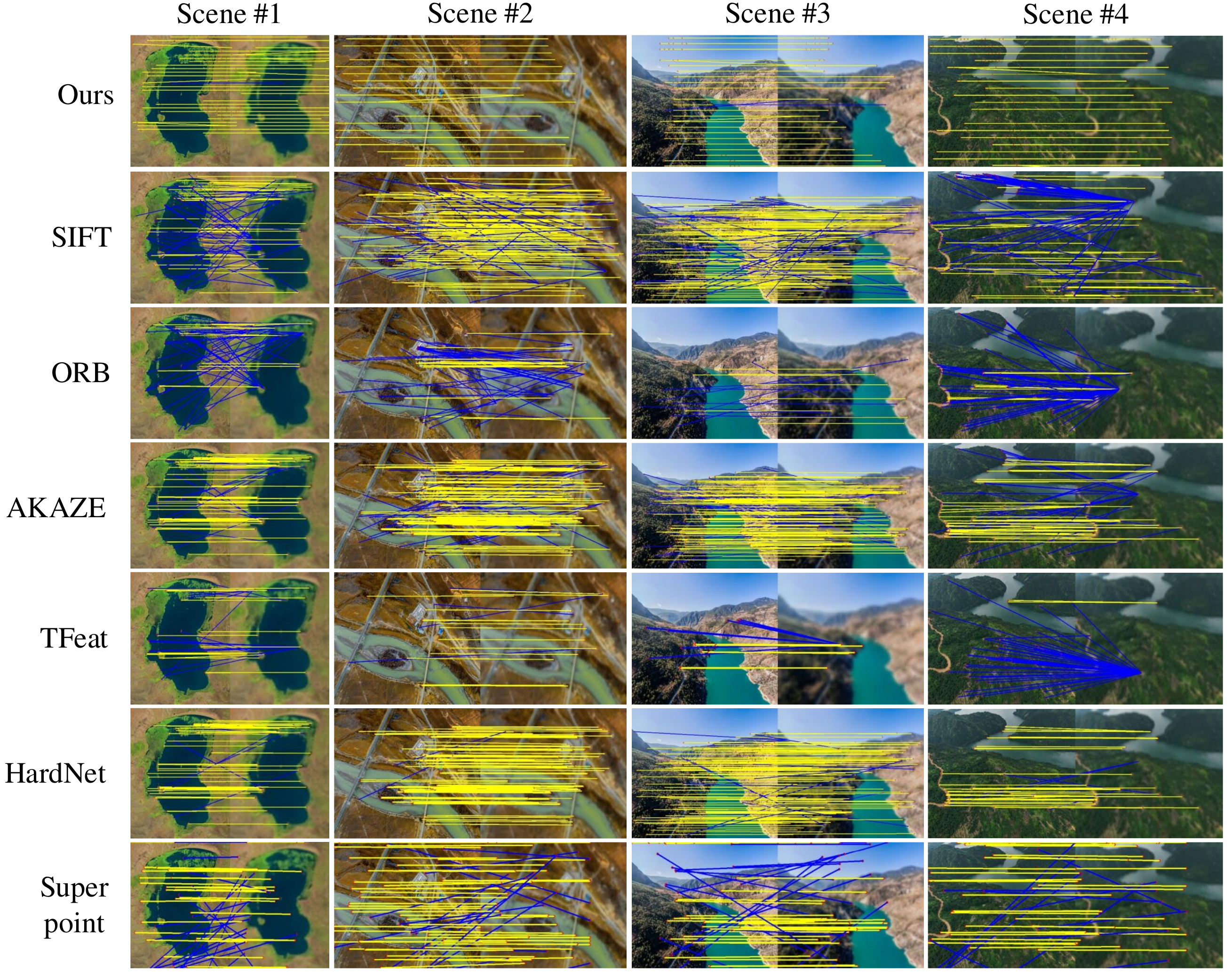}
	\caption{Exemplar feature matching results of different techniques, with the scale difference between two input images being 64. The yellow lines mark the correct matching pairs, and the blue ones mark the wrong matching pairs. The feature matching accuracy of the reported technique outperforms the other algorithms.}
	\label{figure:3}
\end{figure*}

Because feature matching is the most important step in the image registration process, we quantify registration performance using feature point matching accuracy. By denoting the correct feature matching pair as $TP$ and the wrong feature matching pair as $FP$, we define $TPR=TP/(TP+FP)$ as the quantitative metric of feature matching accuracy. The results are shown in Tab. \ref{Tab:01}, with the exemplar visual results of multiscale feature point matching shown in Fig. \ref{figure:3}.

From the results in Tab. \ref{Tab:01}, we can see that our algorithm is superior to the techniques. The performance advantage is more obvious when the scale difference between two input images is larger. The TFeat method obtains the lowest accuracy, which dues to its small network scale and shallow depth. Although it is faster than HardNet and Super-point methods, it cannot extract accurate feature descriptors from the input image blocks, especially when the scales of the input image blocks are different. Among the three traditional feature-matching-based methods, ORB works better when the scale difference is small. However, when the scale differs greatly, the SIFT technique maintains greater adaptability than the other two methods. When the scale difference between the cameras reaches a certain level (such as 256), all the conventional techniques fail to find correct feature matching pairs, while the reported technique still works well with much higher registration accuracy. The visual results in Fig. \ref{figure:3} show the clear comparison of wrong matched feature pairs (denoted by blue connects).

\begin{table}[H]
	\footnotesize
	\centering
	\caption{The feature matching accuracy and running time of different techniques for different scale differences.}
	\begin{tabular}{l|c|c|c|c|c|l}
		\hline
		\diagbox{method}{scale} 
		\multirow{2}{*}{} & \multicolumn{2}{c|}{16} & \multicolumn{2}{c|}{64} & \multicolumn{2}{c}{256} \\
		\hline
		& TPR        & Time      & TPR        & Time      & TPR         & Time      \\
		\hline
		SIFT              & 87.67      & 0.61      & 61.89      & 0.55      & 29.44       & 0.60      \\
		\hline
		ORB               & 90.52      & 0.62      & 66.30      & 0.28      & 10.71       & 0.26      \\
		\hline
		AKAZE             & 89.01      & 0.48      & 51.87      & 0.47      & 12.85       & 0.48      \\
		\hline
		TFeat             & 49.84      & 0.65      & 22.14      & 0.68      & 0           & 0         \\
		\hline
		HardNet           & 91.39      & 2.01      & 74.37      & 1.92      & 31.02       & 1.62      \\
		\hline
		Super-point       & 90.44      & 1.79      & 61.57      & 1.88      & 35.31       & 1.76      \\
		\hline
		Ours              &\textbf{99.89}& \textbf{0.27} &\textbf{ 97.30} &\textbf{ 0.18 } & \textbf{77.06 } &\textbf{ 0.25}    \\
		\hline 
	\end{tabular}
	\label{Tab:01}
\end{table}

%Figure 7 shows some examples of the feature point matching accuracy test results of multi-scale image registration.

%Figure 8 shows some examples of the feature point matching accuracy test results of multi-temporal image registration.

%Figure 9 shows some examples of the feature point matching accuracy test results for multispectral image registration.
%图七
%图八
%图九

\subsection{Adaptability to different white balance parameters}

\begin{figure*}[h]
	\centering
	\includegraphics[width=1.0\linewidth]{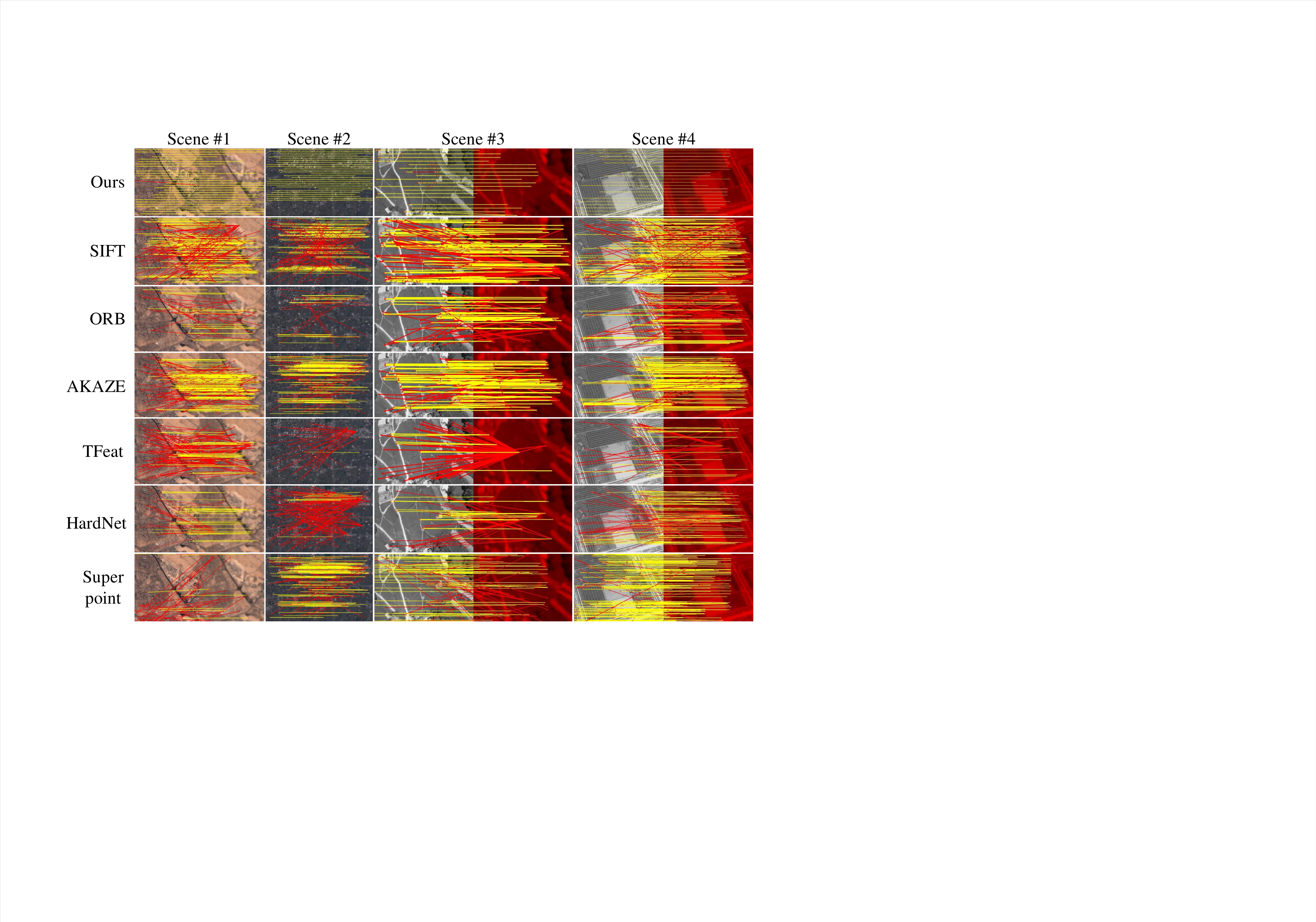}
	\caption{Exemplar feature matching results under different white balance parameters of the input two images, with the scale difference fixed to be 16. The image pairs of $a$ and $b$ are from the OSCD dataset which are acquired at different time. The image pairs of $c$ and $d$ are from the SUIRD dataset, which simulates the situation under large white balance difference. The yellow lines mark the correct matching pairs, and the red ones mark the wrong matching pairs. The feature matching accuracy of the reported technique outperforms the other algorithms.}
	\label{figure:4}
\end{figure*}

In practical applications, two individual cameras may maintain different white balance parameters, which results in spectrum differences of the acquired images. To test the adaptability of the algorithms to different white balance parameters, we synthesize image pairs of different spectrums by color channel separation and run the above algorithms to produce their feature matching accuracy. The quantitative results are shown in Tab. \ref{Tab:02}, with the exemplar visualization results shown in Fig. \ref{figure:4}. We can see that the performance comparison is similar to that in the above section, namely that our algorithm performs best among all the methods. The TFeat technique performed the worst, and our method produced the highest registration accuracy. Among the three traditional methods, the AKAZE algorithm benefits from its nonlinear working space that retains more image detail information when the spectrum of input image changes.

%我们的算法在多光谱多尺度图像配准任务上的性能优于已有算法，且输入图片的尺度相差越大，我们的算法优势越明显。
%表二

\begin{table}[h]
	\footnotesize
	\centering
	\caption{The feature matching accuracy and running time of different techniques for different spectrum under different scales.}
	\begin{tabular}{l|c|c|c|c|c|l}
		\hline
		\diagbox{method}{scale} 
		\multirow{2}{*}{} & \multicolumn{2}{c|}{16} & \multicolumn{2}{c|}{64} & \multicolumn{2}{c}{256} \\
		\hline
		& TPR        & Time      & TPR        & Time      & TPR         & Time      \\
		\hline
		SIFT              & 59.61      & 0.7      & 32.93      & 0.69      & 13.83       & 0.76      \\
		\hline
		ORB               & 43.26      & 0.31      & 22.38      & 0.30      & 5.29       & 0      \\
		\hline
		AKAZE             & 85.17      & 0.48      & 63.18      & 0.45      & 22.04       & 0.47      \\
		\hline
		TFeat             & 9.99      & 0.58      & 3.61      & 0.63      & 1.04           & 0         \\
		\hline
		HardNet           & 73.83      & 1.47      & 55.73      & 0.64      & 20.02       & 0.60      \\
		\hline
		Super-point       & 67.27      & 2.83      & 42.62      & 3.08      & 26.48       & 2.98      \\
		\hline
		Ours              &\textbf{90.71}& \textbf{0.26} &\textbf{ 69.34} &\textbf{ 0.24 } & \textbf{44.62 } &\textbf{ 0.29}    \\
		\hline 
	\end{tabular}
	\label{Tab:02}
\end{table}

\subsection{Experiment}
To validate the effectiveness of the reported agile imaging framework for practical applications, we built a prototype system as shown in Fig. \ref{figure:1}. The reference camera is JAI GO-5000 matched with an LM6HC lens (the focal length is 6mm) to obtain wide-filed images, with an angle of view of $120 \degree$. The dynamic camera is Foxtech Seeker10 enabling two-axis rotation, to obtain high-resolution images with 0.45$mrad$ instantaneous FOV (the focal length reaches up to 49mm). The two cameras are equipped with an NVIDIA TX2 processing platform with Jetpack 3.3 system, which controls camera rotation and image acquisition. The hardware elements are integrated into a 3D-printed shell. The total weight of the entire system is 1181 grams.

\begin{figure*}[h]
	\centering
	\includegraphics[width=1.0\linewidth]{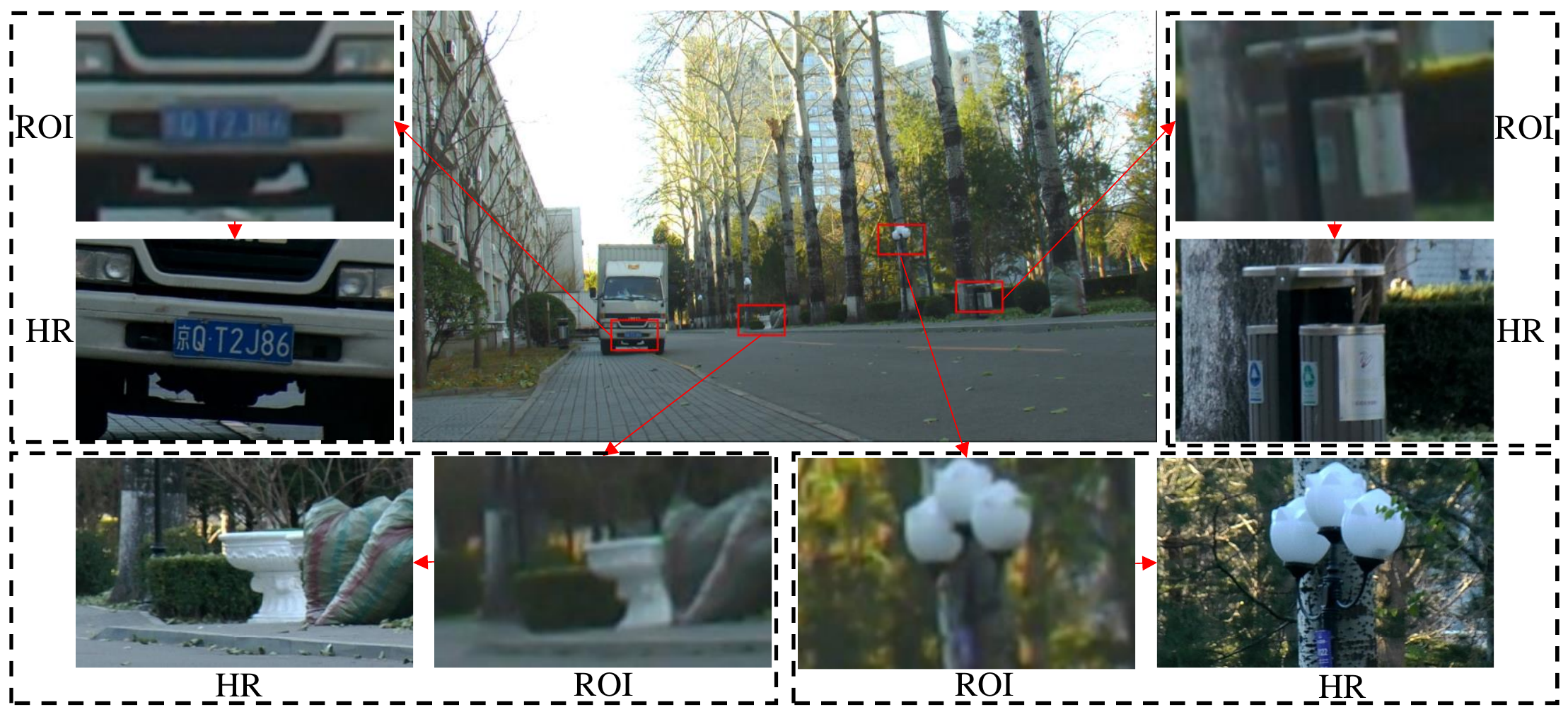}
	\caption{Ground experiment results using the prototypeH system. We employed the prototype to capture images of our campus. The results validate that the setup is effective to capture wide-field images, with selective high resolution on the regions of interests.}
	\label{figure:5}
\end{figure*}

The first experiment is implemented on our campus. We randomly selected a scene and selected four ROI, including car license plates, trash cans, street lights, and flower beds. We used the static reference camera to take wide-field reference images and used the dynamic camera to acquire real-time HR images of the ROI, as shown in Fig. \ref{figure:5}. We can see that the close-ups of the wide-field image are blurred due to limited resolution, and the system is effective to improve the resolution of this ROI. The total data amount is much less than that of the conventional large-scale camera array, relieving storage, transmission, and processing pressure.

\begin{figure}[H]
	\centering
	\includegraphics[width=1\linewidth]{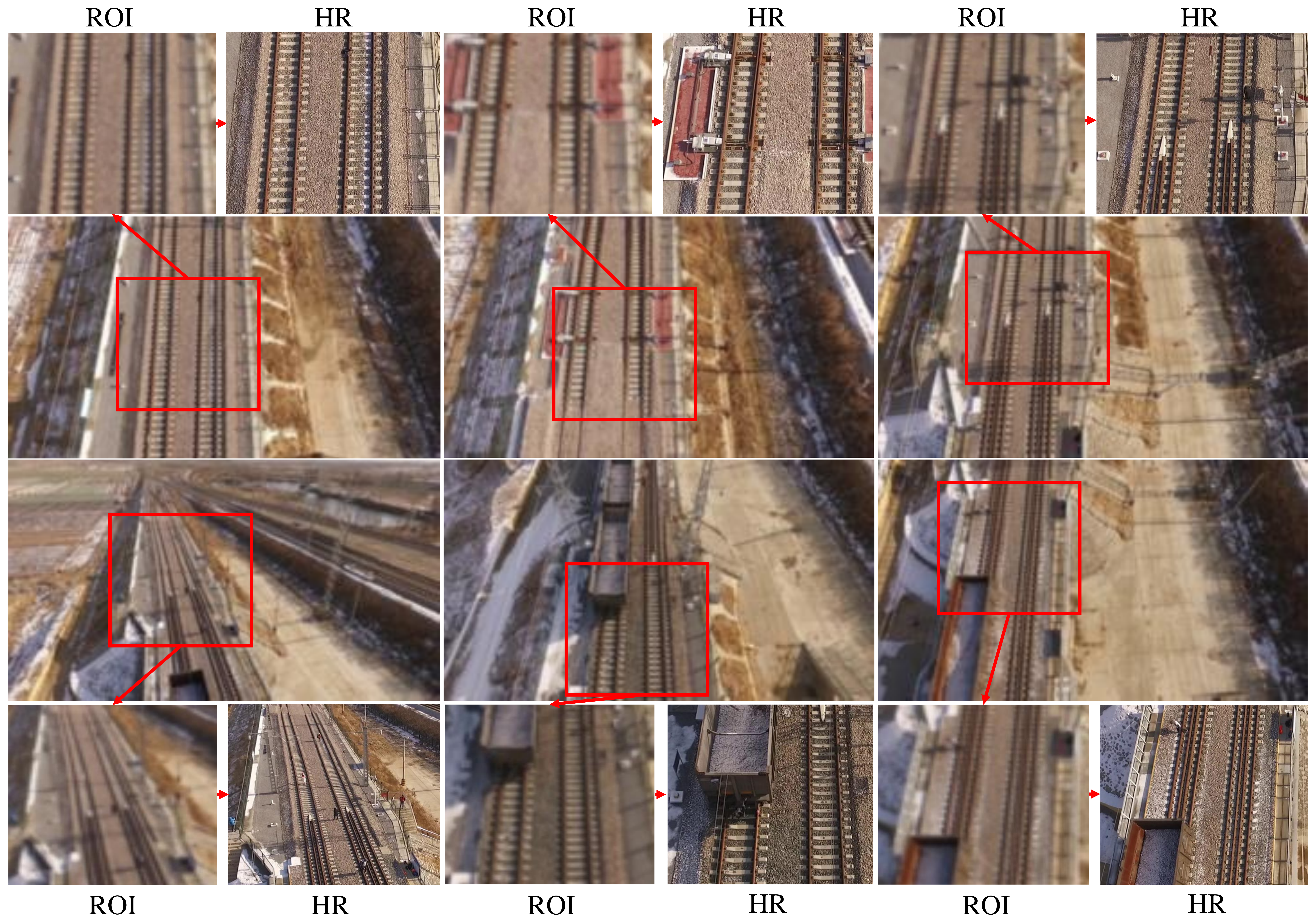}
	\caption{Air-to-ground monitoring experiment results using the prototype setup mounted on a drone. We select a railway as ROI. The reference camera takes wide-field LR images, and the dynamic camera enables continuous HR imaging of the railway using the reported multiscale image registration technique.}
	\label{figure:6}
\end{figure}

We also conducted an air-to-ground monitoring experiment for practical application. We mounted the prototype setup on a DJI M300 RTK drone. The system was equipped with a 4500mAh 11.1V and a 3000mAh 18.5V lithium battery. The former battery provides the 12V DC power required by the two cameras after the voltage regulator module, and the latter one offers the 19V DC power required by the TX2 development platform. The system enables real-time imaging and processing of wide-field HR images during the flight of the drone (10fps).
The drone was flying along the railway, while the wide-field camera ensured that no matter how the drone shakes, HR images of the target railway were always obtained, as shown in Fig. \ref{figure:6}. We can see that the static reference camera takes wide-field low-resolution pictures, and the dynamic high-resolution camera performs continuous real-time tracking and high-resolution imaging of the railway area. This experiment validates the system's wide applications for certain platforms with a limited load.

%静态参考相机拍摄宽视场低分辨率图片，动态高分辨率相机对铁路区域进行连续的实时追踪和高分辨率成像。

\section{Conclusion}

In this work, we reported a novel wide-field and HR imaging architecture based on the sparse prior characteristics of natural scenes. Compared with the conventional gigapixel imaging systems, we reduce the redundant amount of collected data from gigapixel level to megapixel level, with almost the same 120$^{\circ}$ wide field-of-view (FOV) and selective 0.45$mrad$ instantaneous FOV. The system maintains low cost with only two cameras, and the weight is only 1181 grams. To automatically locate ROI in the wide field for continuous and robust HR imaging, we propose an efficient image registration algorithm based on hierarchical convolution features, which overcomes the limitation of existing image registration algorithms that can not handle large scale and white balance parameters between the two input images. We conducted a series of simulations to validate the superiority of our algorithm with higher feature matching accuracy compared to the state-of-the-art, in both cases of large scale difference and white balance difference. Both ground and drone experiments validate the effectiveness of the reported imaging framework, and demonstrate its great potential in practical applications, especially for those with a limited load.

The reported system can be widely extended. First, our algorithm has low requirements on the camera and can be optimized and deployed according to the specific UAV hardware platform, so that it can be widely used on various UAV platforms. And thanks to its extremely fast running speed and small computing power requirements, it can perform real-time HR imaging of ROI on UAV platforms. Second, we can still improve the accuracy and speed of feature point registration by optimizing the feature extraction network structure and using parallel computing. Third, thanks to its extremely fast running speed, our algorithm also has great potential in the field of video object tracking\cite{2006Object} based on feature point matching.

%%%%%%%%%% If using BibTeX:
\bibliography{sample}

%%%%%%%%%% If preparing manually:
% \newcommand{\enquote}[1]{``#1''}

% \bibitem{Zhang:14}
% Y.~Zhang, S.~Qiao, L.~Sun, Q.~W. Shi, W.~Huang, L.~Li, and Z.~Yang,
%   \enquote{Photoinduced active terahertz metamaterials with nanostructured
%   vanadium dioxide film deposited by sol-gel method,}
%   {\protect\JournalTitle{Optics Express}} \textbf{22}, 11070--11078 (2014).

% {Optical Society}, \enquote{{OSA Publishing},}
%   \url{http://www.osapublishing.org}.

% \bibitem{FORSTER2007}
% P.~Forster, V.~Ramaswamy, P.~Artaxo, T.~Bernsten, R.~Betts, D.~Fahey,
%   J.~Haywood, J.~Lean, D.~Lowe, G.~Myhre, J.~Nganga, R.~Prinn, G.~Raga,
%   M.~Schulz, and R.~V. Dorland, \enquote{Changes in atmospheric consituents and
%   in radiative forcing,} in \enquote{Climate Change 2007: The Physical Science
%   Basis. Contribution of Working Group 1 to the Fourth assesment report of
%   Intergovernmental Panel on Climate Change,}  S.~Solomon, D.~Qin, M.~Manning,
%   Z.~Chen, M.~Marquis, K.~B. Averyt, M.~Tignor, and H.~L. Miler, eds.
%   (Cambridge University Press, 2007).

% \end{thebibliography}

\end{document}